\documentclass[final]{cvpr}

\usepackage{times}
\usepackage{epsfig}
\usepackage{graphicx}
\usepackage{amsmath}
\usepackage{amssymb}
\usepackage{flushend}
\usepackage[tight,footnotesize]{subfigure}
\usepackage{multirow}
\usepackage{enumitem}
\usepackage[breaklinks=true,colorlinks]{hyperref}

\begin{document}

%%%%%%%%% TITLE
\title{From Controlled Scenarios to Real-World: Cross-Domain Degradation Pattern Matching for All-in-One Image Restoration}

\author{Junyu Fan$^{1}$, Chuanlin Liao$^{2}$, Yi Lin$^{1}$\thanks{Corresponding author.} \\
$^1$College of Computer Science, Sichuan University, Chengdu, China\\
$^2$The National Key Laboratory of Fundamental Science on Synthetic Vision, \\
Sichuan University, Chengdu, China \\
\centerline{\texttt{\large{junyu.fan@outlook.com, chuanlinliao@stu.scu.edu.cn, yilin@scu.edu.cn}}
}%Institution2\\
}

\pagestyle{empty}  % no page number for the second and the later pages
\maketitle
\thispagestyle{empty}

%%%%%%%%% ABSTRACT
\begin{abstract}
As a fundamental imaging task, All-in-One Image Restoration (AiOIR) aims to achieve image restoration caused by multiple degradation patterns via a single model with unified parameters. Although existing AiOIR approaches obtain promising performance in closed and controlled scenarios, they still suffered from considerable performance reduction in real-world scenarios since the gap of data distributions between the training samples (source domain) and real-world test samples (target domain) can lead inferior degradation awareness ability. To address this issue, a Unified Domain-Adaptive Image Restoration (UDAIR) framework is proposed to effectively achieve AiOIR by leveraging the learned knowledge from source domain to target domain. To improve the degradation identification, a codebook is designed to learn a group of discrete embeddings to denote the degradation patterns, and the cross-sample contrastive learning mechanism is further proposed to capture shared features from different samples of certain degradation. To bridge the data gap, a domain adaptation strategy is proposed to build the feature projection between the source and target domains by dynamically aligning their codebook embeddings, and a correlation alignment-based test-time adaptation mechanism is designed to fine-tune the alignment discrepancies by tightening the degradation embeddings to the corresponding cluster center in the source domain.  Experimental results on 10 open-source datasets demonstrate that UDAIR achieves new state-of-the-art performance for the AiOIR task. Most importantly, the feature cluster validate the degradation identification under unknown conditions, and qualitative comparisons showcase robust generalization to real-world scenarios.
\end{abstract}
%
%%%%%%%%% BODY TEXT
%
%
\section{Introduction}
All things resolved within a single framework is ultimate goal for data-driven tasks, as well as Image Restoration (IR) task in this work.

\begin{figure}[t!]
	\centering
	\includegraphics[width=\linewidth]{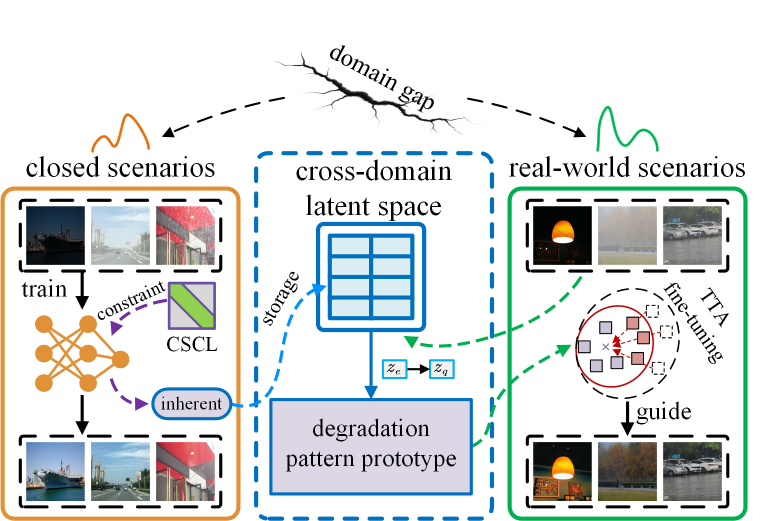}
	\caption{To bridge the gap between closed scenarios and real-world scenarios, the proposed method establishes a cross-domain latent space via a codebook, which stores the shared intrinsic features of degradation patterns constrained through cross-sample contrastive learning. These features serve as prototypes of degradation patterns, effectively leveraging the knowledge and priors learned from closed scenarios. Test-time adaptation is further employed to fine-tune the model for each sample, compensating for errors introduced during cross-domain matching. This strategy mitigates performance decline caused by domain shifts in degradation pattern recognition and guides the network to restore high-quality images from low-quality inputs affected by various degradations.}
	\label{fig1}
\end{figure}

As the fundamental of visual intelligence, the IR task serves as a critical technology to restore degraded visual data into clear and detailed representations, which further provides essential perceptual support for computer vision systems. As the core capability of the artificial intelligence to perceive the physical world, clear and detailed visual images in computer vision system directly determines the potential for humans to break through perceptual boundaries and expand decision-making dimensions.
In past studies, image restoration approaches predominantly focused on individual degradation type in a closed and controlled scenarios, and required different models or weights for different cases (denoising \cite{sdir11}, dehazing \cite{sdir2}, low-light \cite{sdir1}, etc.). 
However, images captured in real-world is with diverse and complex degradation patterns, which presents a data gap from the ideal case in controlled scenarios \cite{domain3}, each images is with only a single degradation. In this context, to achieve the IR task with multiple degradations, a set of models are required to separably built and trained for each degradation, which reduce the generalizability and robustness in practical applications. 
For instance, in the airport surface surveillance scenarios under hazing, raining, and low light conditions, different IR models are deployed and sequentially performed to obtain a clear and detailed monitoring image to support the downstream tasks, which not only burdens the total resource requirements and system latency but also causes unexpected cascaded failures (each IR model has their own unique task objective).
Therefore, an All-in-One Image Restoration (AiOIR) framework has emerged as a promising solution to tackle the images with multiple degradation patterns using only a single model with unified parameters \cite{aioirsurvey, domain1}. 
The AiOIR is expected to provide a general solution for degraded image restoration in complex environments, such as autonomous driving \cite{app1}, unmanned aerial vehicle \cite{app2}, underwater robotics \cite{app4}, and other applications \cite{app3, app5}.

Currently, image restoration methods can be divided into single degradation image restoration (SDIR), multiple degradation image restoration (MDIR), and All-in-One image restoration. 
The SDIR methods, such as \cite{sdir6, sdir9, sdir10}, are typically designed for specific tasks and are particularly effective at handling known degradation types. However, these methods suffered from poor performance facing unknown degradation patterns. 
The MDIR methods, such as \cite{mdir1, mdir2, mdir3, mdir4}, attempt to develop a unified framework to flexibly cope with various degradation with degradation-related model weights. Although MDIR methods achieve higher generalizability for IR tasks, they still require the prior degradation types to perform pre-selected IR operations in practical applications. 
Recently, AiOIR methods, such as \cite{adair, captnet, mioir, daclip}, are proposed to simultaneously address multiple unknown degradation tasks using a single framework with unified parameter weights, thereby providing both generality and practicality in real-world tasks.

In general, the key to AiOIR approaches depends on the ability to effectively identify degradation patterns that are usually unknown and cannot be determined in advance. In addition, in real-world scenarios, data gap of different domains inevitably deteriorate the performance of existing methods with degradation prior assumptions from close scenarios datasets \cite{aioirsurvey, domain1}, indicating the dual challenges bias in identifying degradation patterns and domain distribution shift \cite{domain2}, as shown below:

\begin{itemize}
    \item In recent works, prompt-based learning \cite{promptir}, contrastive learning \cite{airnet}, and multi-modal \cite{multimodal} were proposed and improved to enhance the representation capabilities of complex degradations. However, the limitation of these methods to single-sample optimization results in learning sample-wise features with irrelevant to degradation, thereby failing to model the shared characteristics of degradation patterns across samples. These limitations lead to insufficient identify accuracy and inability to effectively capture degradation features, ultimately impairing the restoration performance.
    
    \item Facing the distribution shift among different image domains, it is hard to directly transfer the learned representations from the source domain data to the target domain data, which further fails to capture degradation patterns and corresponding degradation features. Actually, existing methods can only obtain desired performance in training closed scenarios (source domain), while suffering from poor generalization ability in environments with different feature distributions (target domain). Obviously, a single dataset can not cover the images with all the degradation scenarios, and the data collection is time-consuming and labor-intensive. In this context, it is extremely hard to build a comprehensive dataset to tackle the distribution shift issue. 
\end{itemize}

To address the aforementioned issues, a Unified Domain-Adaptive Image Restoration framework (UDAIR) is proposed to effectively utilize the knowledge and priors learned from the source domain during inference with real-scenario data, as shown in Fig.\ref{fig1}. The core idea of the proposed UDAIR redefines the AiOIR task,i.e., solving complex inter-domain differences by building and matching cross-domain degradation pattern prototypes.

To enhance the degradation perception identification across different data domains, a Cross-Sample Contrastive Learning (CSCL) mechanism is proposed to constrain the model to explore the shared intrinsic representations of a certain degradation pattern across diverse images. CSCL group the degradation features of multiple samples by task and apply random permutation to generate positive sample pairs with the same degradation pattern but exhibit different degradation manifestations.     
To this end, a codebook-based Degradation Aware and Analysis Module (DAAM) is designed to establish cross-domain latent space to store degradation pattern prototypes by learning shared intrinsic priors of degradation pattern rather than sample-wise local features unrelated to degradation, which is expected to be applicable in identifying this degradation in unseen data domains. The codebook is serves to match the prototype that most closely corresponds to the degradation pattern of the current sample, leveraging the intrinsic properties of degradation prototypes to build a bridge across domains and alleviate the issue of data distribution shift.

Considering the inherent distribution discrepancies of images between source and target domain, in this work, a correlation alignment Test-Time Adaptation (TTA) \cite{tent, tta2} strategy is proposed to perform the online re-adjustment via progressive alignment based on the sample-wise features, which benefits to mitigate matching errors caused by inter-domain distribution discrepancies. Existing work \cite{ttamethod} typically fine-tunes the IR model depending only on the target domain data via TTA, with insufficient attention on the learned prior of degradation types in the source domain, which leads to catastrophic forgetting and further reduce solution performance \cite{ttaissue2}. In general, the IR model optimized on the source domain data can generate informative features to support the identification of degradation patterns, which extracts intrinsic and generalizable restoration prior to denote a shared latent representation of degradation patterns in the target domain. Unlike the direct domain alignment strategies in existing TTA, for each image degradation, by regarding the cluster center of the learned degradation features in the source domain as the anchor, the degradation features obtained from target domain data are pulled to their corresponding cluster center by learnable transformation and correlation alignment, which is to mitigate the distribution discrepancies by formulating a shared degradation representations between source and target domain.

To validate the proposed UDAIR model, a total of 5 common IR tasks, concerning denoising, dehazing, deraining, low-light image enhancement (LLIE), and underwater image enhancement (UIE), are selected to conduct comparison experiments. For each task, two different datasets serve as the source and target domain, respectively. Recent competitive baselines are also applied to evaluate the model performance in terms of both common IR metrics and the specific task-related measurements. Extensive experimental results demonstrate that the UDAIR outperforms other baselines and achieves new state-of-the-art performance for the AiOIR task. Most importantly, the visualizations of feature cluster validate the degradation identification under unknown conditions, and qualitative comparisons showcase robust generalization to real-world scenarios, which finally supports the innovative motivations in this work.

\begin{itemize}
    \item A Unified Domain-Adaptive Image Restoration framework is proposed to effectively mitigate performance reduction caused by distribution discrepancies, which leverages learned knowledge and priors from the source domain to construct a cross-domain latent space to match degradation patterns in the target domain. The UDAIR model enhances both the degradation pattern identification and restoration performance in real-world scenarios.
    
    \item A Cross-Sample Contrastive Learning mechanism is proposed to learn informative representations for intrinsic nature of image degradation by  constructing sample pairs from different samples with certain degradation patterns, which is also able to suppress the interference features to further enhance the identification capacity of the degradation patterns. 
    
    \item A codebook-based domain adaptation strategy is proposed to establish a shared cross-domain feature space by preserving learned prototypes of degradation patterns, based on which the TTA is combined to perform the dynamic fine-tuning for the UDAIR model. The domain adaptation is to address distribution shifts of the cross-domain samples, and the decreasing Kullback–Leibler (KL) divergence further demonstrates the effectiveness of the proposed strategy.
    
    \item Extensive experimental results demonstrate that the proposed model achieves new state-of-the-art performance in an AiOIR way for all five tasks on both 10 open-source closed and real-world scenario datasets, in terms of all the proposed metrics. The in-depth analysis also supports the research motivation in this work and validates the potential for real-world applications.
\end{itemize}

\begin{figure*}[t!]
	\centering
	\includegraphics[width=0.95\linewidth]{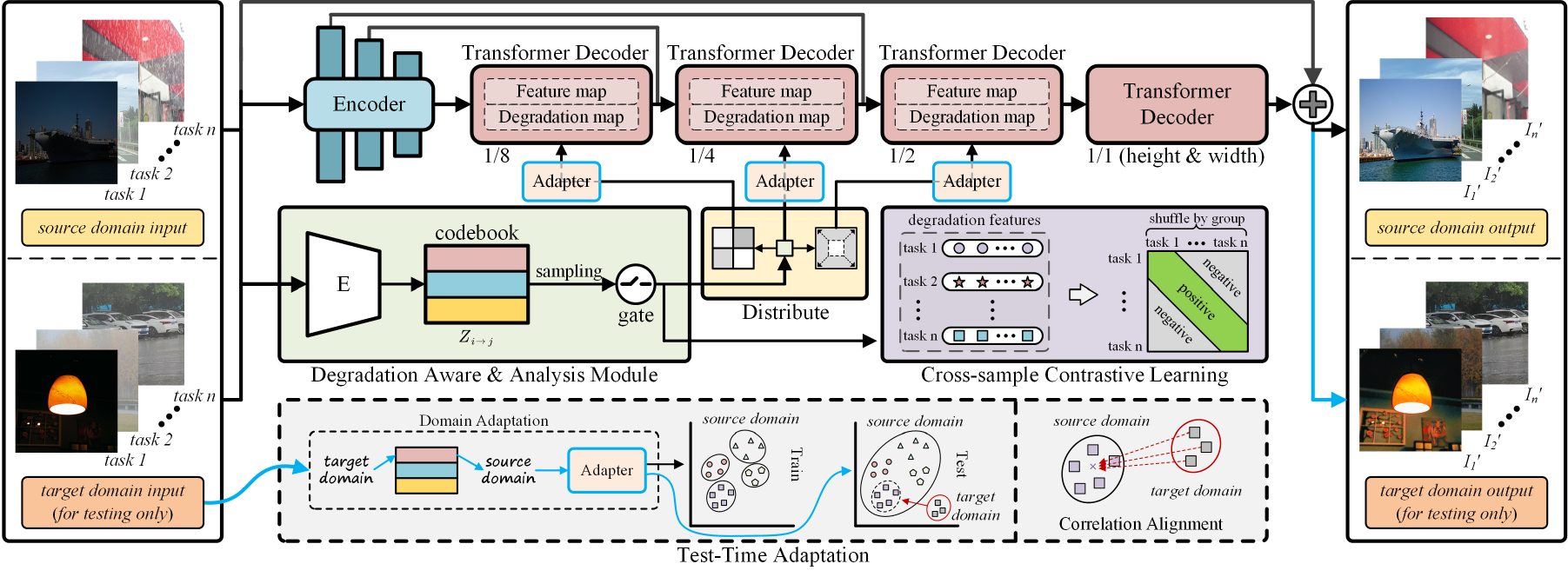}
	\caption{Overview of the proposed framework. The blue pipeline represent the dynamic domain adaptation strategy activated exclusively during target domain inference.}
	\label{net}
\end{figure*}

\section{Method}
\subsection{Motivation} \label{sec5.1}
Degradations are often diverse and complex, and current AiOIR methods are usually trained on relatively closed and controlled datasets. Due to feature distribution shifts between the source and target domains in real-world applications, the model shows weak generalization on the target domain. Specifically, distribution shifts hinder the direct applicability of the learned representations to the target domain data, resulting in a significant decline in restoration performance.

To address this challenge, a Unified Domain-Adaptive Image Restoration framework is proposed to effectively leverage the knowledge learned from source domain to target domain, as shown in Fig.\ref{net}. 
For cross-domain degradation perception, a Cross-Sample Contrastive Learning mechanism is proposed to constrain models to capturing and learning the shared intrinsic features of degradation patterns. Specifically, the extracted image components are grouped by task, and random permutation is applied to generate positive sample pairs that share the same semantics but exhibit different degradation patterns. Through cross-sample contrastive learning, the model is encouraged to autonomously capture and analyze the representation of degradation features without relying on references or labels. Compared to constructing positive pairs from different regions within the same image, this strategy covers a more diverse degradation representations. This allows the model to better capture the commonalities of degradation features, focus on the essential differences among degradation patterns, and ultimately improve its generalization ability across various degradation scenarios.

A domain adaptation strategy is considered to address data distribution shifts via codebook and TTA. The codebook establishes a cross-domain latent space storing degradation pattern prototypes, enabling the model to leverage learned knowledge and priors from the source domain. However, due to domain discrepancies, this approach inherently introduces discretization errors and information loss during prototype matching. To mitigate these alignment inaccuracies, TTA dynamically adjusts model parameters based on local characteristics of test samples, thereby alleviating generalization errors caused by inter-domain distribution shifts.

\subsection{Backbone} \label{sec5.2}
In the backbone, the Vision Transformer \cite{vit} is primarily used as the core structure for both the encoder and decoder, with each layer consisting of $L$ Transformer blocks.

Since the degradation in the input image is unknown, the captured degradation features are leveraged to guide the decoder in restoring images with various degradation types. These degradation features serve as an abstract representation of specific degradation types, capturing the diverse patterns caused by complex factors in the input image and encoding low-dimensional semantic information into compact feature vectors. Specifically, before each decoder layer, features from two sources are concatenated: one part consists of features passed from the encoder, which contain the basic structural information of the input image; the other part comprises the degradation features from the Degradation Aware and Analysis Module (DAAM). Additionally, before inputting the degradation features, a Domain Adaptation Module (DAM) is designed to further align the target domain with the source domain during TTA (the DAM will be explained in a later subsection).

\subsection{Degradation Aware and Analysis} \label{sec5.3}
\textbf{Degradation awareness.} A convolutional neural network as the feature extractor, with the primary task of progressively downsampling high-dimensional input features and extracting key information to generate a low-dimensional latent representation. With the help of a codebook, complex degradation patterns can be quantized into a set of learnable representations. Specifically, each vector at a spatial location in the low-dimensional feature is treated as an individual encoding unit and is flattened into a two-dimensional matrix, where each row corresponds to a continuous feature vector $z_e$ that is output by the encoder at a given spatial location. Meanwhile, the predefined codebook is an embedding matrix of size $[K,D]$, where $K$ represents the total number of discrete codes and $D$ is the length of each discrete code. For each flattened vector $z_e$, the Euclidean distance to each vector $e_k(k=1,2,…,K)$ in the codebook is computed as:
\begin{equation} \label{eq1}
    d\left( z_e,e_k \right) =\left\| z_e \right. -\left. e_k \right\| _2
\end{equation}

For each $z_e$,  a quantization operation is performed by assigning it to the discrete code $k^*$ with the smallest distance, mapping the continuous feature to a discrete code value:
\begin{equation} \label{eq2}
	k^*=\mathrm{arg}\min _k\left\| z_e \right. -\left. e_k \right\| _2
\end{equation}

After obtaining the discrete index corresponding to each $z_e$, the associated embedding vector, denoted as $z_q$, is retrieved from the codebook. Finally, $z_e$ is replaced by its nearest neighbor codebook vector:
\begin{equation} \label{eq3}
    z_q=e_{k^*}
\end{equation}

This mechanism effectively achieves inter-domain mapping by constructing a cross-domain latent space that stores prototypes capturing the shared inherent nature of degradation patterns and exhibiting strong cross-domain generalizability. Such architecture mitigates performance degradation caused by inter-domain distribution discrepancies, thereby enhancing both the accuracy of degradation pattern capture and restoration performance in real-world scenarios.

\begin{figure}[t!]
	\centering
	\includegraphics[width=0.9\linewidth]{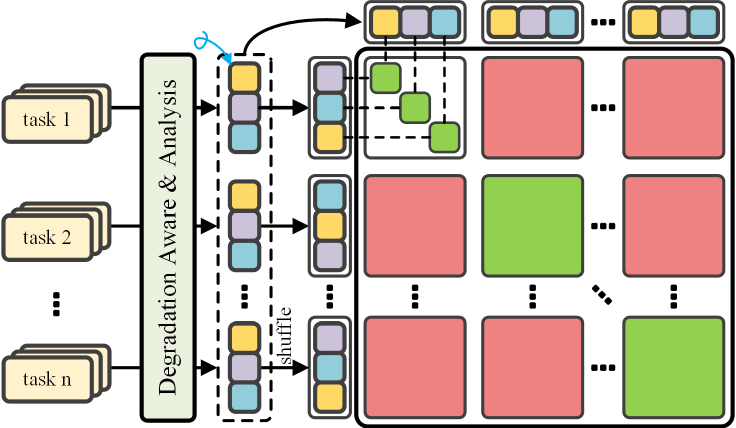}
	\caption{Schematic diagram of Cross-Sample Contrastive Learning.}
	\label{cscl}
\end{figure}

The gating mechanism filters these discrete features and assigns weights to them, selecting the most representative feature information. After gating, the model can focus on features that are especially salient in expressing similar degradation information and that exhibit high discriminability and robustness. This not only reduces redundancy but also enhances the effectiveness of subsequent contrastive learning stages.

Extract a global embedding $E$ for each channel:
\begin{equation} \label{eq4}
    E=\lambda \sqrt{\sum_{h=1}^H{\sum_{w=1}^W{x_{b,c,h,w}^{2}+\epsilon}}}
\end{equation}
where $b$ represents the sample, $c$ represents the channel, $x$ denotes the input feature, $\lambda$ is a learnable scaling factor, and $\epsilon$ is a small constant to prevent division by zero. Then, this embedding is used to compute a normalization factor $N$:
\begin{equation} \label{eq5}
	N=\sqrt{\frac{1}{C}\sum_{c=1}^C{E^2+\epsilon}}
\end{equation}

The gating coefficient $g$ is obtained by applying a $tanh$ mapping function to the global embedding $E$ and the normalization factor $N$. This coefficient is then multiplied channel-wise with the original feature to produce the final output $F_g$:
\begin{equation} \label{eq6}
	g=1+\tanh \!\:\left( \gamma _c \cdot \frac{E}{N}+\beta _c \right) 
\end{equation}
\begin{equation} \label{eq7}
	F_g=x\cdot g
\end{equation}
where $\gamma_c$ and $\beta_c$ are learnable parameters corresponding to scaling and bias, respectively.

\begin{table}[h]
	\centering
	\resizebox{\linewidth}{!}{
		\begin{tabular}{p{\linewidth}}
			\hline
			\textbf{Algorithm of CSCL}  \\
			\hline
			\textbf{Input: }\\
			~~~~$\mathrm{F_{d}}$: Degradation features of shape [batch\_size, features]\\
			~~~~$\mathrm{N_{t}}$: Number of tasks\\
			~~~~$\mathrm{N_{s}}$: Number of samples per task\\
			~~~~CL: Contrastive loss function
			 \\
			\textbf{Output: }\\
			~~~~loss: Computed loss value \\
			\\
			1: Reshape to task-wise mini-batches:\\
			~~~~$F_{t} = reshape(\mathrm{F_{d}},~[\mathrm{N_{t}},~\mathrm{N_{s}},~\mathrm{features}])$ \\
			\\
			
			2: Aggregate original features by task: \\
			~~~~$F_{g} = reshape(F_{t},~[\mathrm{N_{t}},~\mathrm{N_{s}} * \mathrm{features}])$ \\

			\\
			3: Shuffle samples within each task: \\
			~~~~Initialize an empty list: $F_{s}$ = [ ]\\
			~~~~$for~~i = 0~~to~~\mathrm{N_{t}} - 1:$ \\
			~~~~~~~~$I_{p} = random\_permutation(samples)$ \\
			~~~~~~~~$shuffled = F_{t}[i,~I_{p},~:]$ \\
			~~~~~~~~$append~~shuffled~~to~~F_{s}$ \\
			~~~~$end~for$ \\
			\\
			
			4: Aggregate shuffled features by task: \\
			~~~~$F_{s} = reshape(cat(*[F_{s}]),~[\mathrm{N_{t}},~\mathrm{N_{s}} * \mathrm{features}])$ \\
			\\
			
			5: Compute contrastive loss: \\
			~~~~$loss = \mathrm{CL}(F_{g},~F_{s})$ \\
			\\
			
			6. Return $loss$
			\\
			
			\hline
		\end{tabular}%
	}
	\label{algorithm}%
\end{table}

\textbf{Cross-sample contrastive learning.} Contrastive learning can automatically learn and capture various degradation features from large amounts of data. Its core idea is to pull similar samples closer in the feature space while pushing dissimilar samples apart. In contrastive learning, positive sample pairs are formed by samples of the same degradation type, whereas negative sample pairs consist of samples of different degradation types. To introduce more variability in the positive sample pairs, features from samples with the same degradation pattern are group and randomly permute, enabling the model to learn common and stable cross-sample feature representations. The diagram of CSCL as shown in Fig.\ref{cscl}, and the algorithm of CSCL is presented in Algorithm. Let the set of sample features for a certain degradation type $d$ be denoted as 
\begin{equation} \label{eq8}
	S_d=\{s_1,s_2,\cdots ,s_n\}
\end{equation}

For the set $S_d$ of sample features, a random permutation function $\pi$ is applied to obtain a permuted order:
\begin{equation} \label{eq9}
	S_{d}^{\pi}=\{s_{\pi (1)},s_{\pi (2)},\cdots ,s_{\pi (n)}\}
\end{equation}

Then, construct positive sample pairs
\begin{equation} \label{eq10}
	\mathcal{P} =\{S_d,S_{d}^{\pi}\}
\end{equation}

and each negative sample pair is given by
\begin{equation} \label{eq11}
	\mathcal{N} =\{(S_d,S_{d^{\mathrm{'}}}^{\pi})|d\ne d^{\mathrm{'}}\}
\end{equation}

This approach, which does not rely on local features from a single image, encourages the model to focus on more robust, abstract, and general semantic information, thereby enhancing its generalization across different scenarios or data distributions. By combining with a dataloader that randomly selects samples, positive and negative sample pairs can nearly cover the entire dataset. Such diversity prevents the model from learning spurious correlations due to fixed data ordering, allowing the contrastive learning process to concentrate on the semantic consistency between positive sample pairs rather than being constrained by a specific order or local structure. The cross-sample contrastive learning loss is computed as follows:
\begin{small}
    \begin{equation} \label{eq12}
	\mathcal{L}_{CSCL}=-\frac{1}{N}\sum\nolimits_{i=1}^N{log\frac{\exp(\mathrm{sim}{{\left( \mathcal{P} _i \right)}/{\mathrm{\tau}}})}{\sum\nolimits_{j=1}^N{\exp( \mathrm{sim}{{\left(\mathcal{N}_{i,j} \right)}/{\mathrm{\tau}}})}}}
\end{equation}
\end{small}

where $sim$ denotes cosine similarity; $\tau$ is a tuning parameter used to adjust the smoothness of the distribution; and $N$ represents the number of degradation types.

With this strategy, the model can autonomously capture and differentiate the degradation-relevant core features without relying on external references or labels, thereby avoiding limitations imposed by local information within a single image. This leads to more robust and efficient learning of degradation features.

\begin{figure}[t!]
	\centering
	\includegraphics[width=0.9\linewidth]{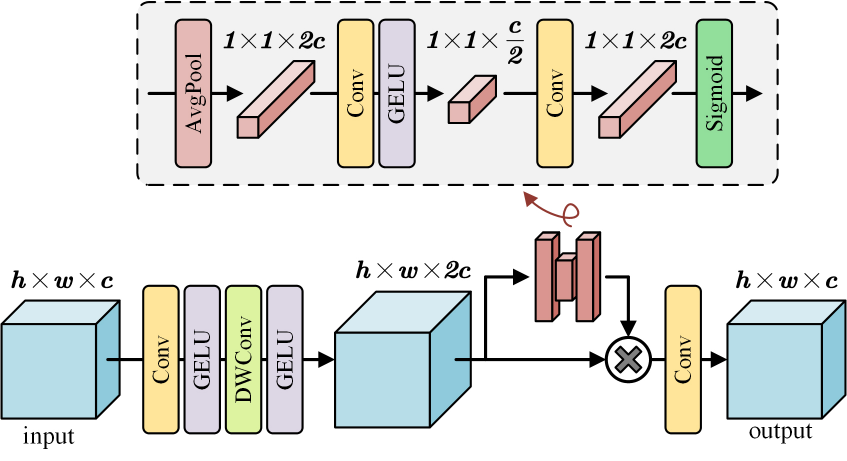}
	\caption{Schematic diagram of Domain Adaptation Module.}
	\label{dam}
\end{figure}

\subsection{Domain Adaptation} \label{sec5.4}
Source domain data is typically synthetic or captured in specially constructed scenes. However, in practical applications, the target domain data may exhibit feature distributions that are completely different from those of the source domain, causing models trained on source domain data to have poor performance when generalized to the target domain. Therefore, a domain adaptation strategy is needed to mitigate the performance degradation caused by domain distribution differences.

Although the codebook has established a cross-domain latent space to match degradation prototypes across different domains, quantization errors and loss of fine-grained information are inevitable. Therefore, TTA and DAM are introduced for further alignment and correction. Specifically, a DAM is construct to establishes a mapping from the current degradation features to anchor points. These anchor points are set as the cluster centers of each degradation pattern in the source domain feature space. When using source domain data, the DAM remains inactive and does not participate in gradient computation or weight updates, thereby preventing interference with the established feature representations. In contrast, when using target domain data, the DAM is activated, gradients are computed, and weights are updated to dynamically correct the feature distribution.

DAM, as shown in Fig.\ref{dam}, expands the channels (set to twice the original in experiment) of the input degradation features using a $1\times1$ convolution operation and further captures local spatial information with a $3\times3$ depthwise separable convolution operation. This not only enhances the expressive capability and reduces computational complexity, but is also particularly critical for recovering fine-grained features in the degraded data. The specific implementation is as follows:
\begin{equation} \label{eq13}
    F_{adapt}=\mathrm{\sigma (DWC(\sigma (Conv}_{1\times 1}(F_d))))
\end{equation}
where $F_d$ represents the degradation features, $Conv_{1\times1}$ represents the convolution operation with a $1\times1$ kernel, DWC represents the $3\times3$ depthwise separable convolution, and $\sigma$ is the GELU activation function.

By embedding the Squeeze-and-Excitation (SE) module, DAM aggregates important information from each channel on a global scale, enabling the channel weights to be adaptively adjusted. This helps to emphasize the feature responses that are most crucial for mapping to the anchor points. The specific implementation is as follows:
\begin{small}
    \begin{equation} \label{eq14}
    F_{se}=F_{adapt}*\delta (\mathrm{Conv}(\sigma (\mathrm{Conv}(\mathrm{AvgPool}(F_{adapt})))))
\end{equation}
\end{small}
where $\delta$ is the Sigmoid activation function.

Finally, the precessed feature map is restored to its original dimensions by a $1\times1$ convolution operation.
\begin{equation} \label{eq15}
	F_{dam}=\mathrm{Conv}_{1\times 1}(F_{se})
\end{equation}

In this process, TTA is responsible for updating the weights of the DAM by using the CORAL \cite{coral} loss to constrain the alignment between the target and source domain feature distributions. Specifically, when testing with target domain data, the DAM is activated during both fine-tuning and inference stages. During this process, the weights of the DAM are selected and gradients are enabled, allowing the CORAL loss to be computed between the output degradation features from DAM and the anchor. This loss is then used to update the weights of DAM.

The core idea of the CORAL loss function is to measure the distributional discrepancy between the source and target domain features by computing the difference between their covariance matrices. Given the source domain feature matrix $S$ and the target domain feature matrix $T$, their covariance matrices are computed as follows:
\begin{equation} \label{eq16}
	C_s=\frac{S^{\mathrm{T}}S-\frac{1}{n_s}\left( S^{\mathrm{T}}1_{n_s} \right) \left( 1_{n_s}^{\mathrm{T}}S \right)}{n_s-1}
\end{equation}
\begin{equation} \label{eq17}
	C_t=\frac{T^{\mathrm{T}}T-\frac{1}{n_t}\left( T^{\mathrm{T}}1_{n_t} \right) \left( 1_{n_t}^{\mathrm{T}}T \right)}{n_t-1}
\end{equation}
where $n$ denotes the number of samples, and $1_n$ represents a $1\times n$ vector with all elements equal to one. After obtaining the covariance matrices for the source and target domain features, the Frobenius norm of the difference between these matrices is calculated by
\begin{equation} \label{eq18}
	\left\| C_s \right. -\left. C_t \right\| _F=\sqrt{\sum_{i,j}{\left( C_{s}^{\left( i,j \right)}-C_{t}^{\left( i.j \right)} \right) ^2}}
\end{equation}

Finally, the CORAL loss is obtained by normalizing this value:
\begin{equation} \label{eq19}
	\mathcal{L} _{CORAL}=\frac{1}{4d^2}\left\| C_s \right. -\left. C_t \right\| _F^2
\end{equation}
where $d$ is the dimensionality of the feature representations.
% Set cluster center of degradation features is $\{c_k\}_{k=1}^K$, the target domain sample features is $f_t$, the optimization objective of TTA can be defined as
% \begin{equation} \label{eq20}
% 	\mathop{\mathrm{min}}\limits_{\theta}\sum_{i=1}^N||TTA_\theta(f_t^{(i)})-c_{k^*}||_2^2
% \end{equation}

\subsection{Loss Function}
During training, the network constrained primarily with the MAE loss and the CSCL loss:
\begin{equation} \label{eq20}
	\mathcal{L} _{total}=\alpha \mathcal{L} _{MAE}+\beta \mathcal{L} _{CSCL}
\end{equation}
where $\alpha$ and $\beta$ are the weights for the respective losses and are set to 1 and 0.2 to balance them on the same scale in this experiment. The $\mathcal{L} _{MAE}$ is defined as follows:
\begin{equation} \label{eq21}
	\mathcal{L} _{MAE}=\frac{\sum\nolimits_{i=1}^n{|f\left( x_i \right) -y_i|}}{n}
\end{equation}
where $f(x_i)$ and $y_i$ denote the prediction and its corresponding ground truth for the $i$-th sample, respectively, and $n$ is the total number of samples.

\section{Results}
subsection{Task Overview} \label{sec3.1}
Given a degraded image $y \in \mathbb{R}^{H \times W \times C}$, the corresponding clean image $x \in \mathbb{R}^{H \times W \times C}$, the degradation process can be modeled as
\begin{equation}
    y = \mathcal{D}(x; d)
\end{equation}
where $\mathcal{D}$ is the degradation function, $d$ is the degradation type.

For SDIR and MDIR methods, separate models are required to be trained for specific degradation types or with pre-selected degradation type to restore clean image $\hat{x}$, as in:
\begin{equation}
    \hat{x_i}=\mathcal{N}_{\theta_i}(y;d_i)
\end{equation}
where $\mathcal{N}$ is the trainable neural network, $\theta_i$ is optimized only for a certain degradation pattern $d_i$. Therefore, $k$ sets of model weights $\{\theta_1, \theta_2, ..., \theta_k\}$ are optimized with different training samples for each degenerate pattern $\{d_1, d_2, ..., d_k\}$.

In this work, AiOIR recovers the clean image with degradation awareness and adaptive restoration by a single model weight, in which the model can automatically detect the degradation patterns from $y$ by implicit degradation perception:
\begin{equation}
    \hat{x}=\mathcal{N}_\theta(y)=\mathcal{R}(y,\mathcal{A}(y;\theta_a); \theta_r)
\end{equation}
where $\mathcal{R}$ is the restoration operation, $\mathcal{A}$ is the degradation identification operation, $\theta=\{ \theta_a, \theta_r \}$ is unique and independent of degradation pattern $d$, $\theta_a$ denotes the weights of degradation identification, and $\theta_r$ denotes the weights of image restoration.

In addition, the generalization challenge of the AiOIR task in real-world scenarios are formulated as a domain adaptation problem. During the model inference for target domain samples, the proposed UDAIR framework introduces TTA to optimize the domain adaptation module, as in:
\begin{equation}
    \theta'_{da}=\theta_{da}-\eta \cdot \nabla_{\theta_{da}}\mathcal{L}_{TTA}
\end{equation}
where $\theta_{da}$ is the weights of Domain Adaptation Module, which are activated and updated only during the testing procedure on target domain samples. $\eta$ denotes the learning rate, and $\nabla$ denotes gradient operator.

Finally, the test procedure in the target domain can be described as:
\begin{equation}
    \hat{x}=\mathcal{N}_\theta(y)=\mathcal{R}(y,\mathcal{A}(y;\theta_a;\mathcal{T}(\theta_{da})); \theta_r)
\end{equation}
where $\mathcal{T}$ is the TTA mechanism.

\subsection{Dataset and Data Preprocessing} \label{sec3.2}
In this work, four common real-world imaging challenges (sensor noise, aerosol scattering, rain streak occlusion, and low-light) are focused to validate the proposed model, in which each corresponds to a distinct physical degradation process. In general, the mentioned challenges are inherently caused by the interplay of imaging hardware limitation (e.g., noise) and complex environmental conditions (e.g., haze, rain, low-light), which concerns the denoising, dehazing, deraining, and low-light image enhancement tasks for critical computer vision systems.
In addition, the underwater image enhancement task is also considered to explore the model ability to complex degradations. In the underwater environment, most atmospheric degradations result from particle scattering and absorption, in which the noticeable color attenuation, non-uniform lighting, and low contrast are also presented due to the water turbidity. The UIE task requires the IR model not only to remove occlusions but also to perform accurate spectral and color corrections. It is believed that the mentioned task design can provide a more demanding challenge to formulate a solid evaluation across diverse degradations.

For each mentioned IR task, two independent datasets are selected to represent the data distribution in the source and target domain, respectively, which further explores the model capacity to address the distribution shifts among different domains (datasets). In this context, a total of 10 open-source datasets are determined to conduct the comparison experiments, as shown below:

Source domain datasets: SIDD \cite{sidd} (Denoising), OTS \cite{reside} (Dehazing), RealRain‑1k \cite{realrain} (Deraining), LOL \cite{lol} (LLIE), UIEB \cite{uieb} (UIE).

Target domain datasets: Polyu \cite{polyu} (Denoising), URHI \cite{reside} (Dehazing), LHP‑Rain \cite{lhp} (Deraining), LIME \cite{lime} (LLIE), UFO‑120 \cite{ufo} (UIE).

In general, the source domain datasets are selected to approximate real-world conditions by either synthetic or real collected images, while the samples in target domain datasets are collected from real scenes to imitate the distribution shifts, which is expected to formulate a robust controllable domain mapping and alignment.

Since the proposed UDAIR model is trained to perform the AiOIR on all the five IR tasks, it is required to organize a balanced dataset across different degradation patterns to enhance the model training and convergency. To this end, a random selection strategy is applied to sample images from the raw dataset, which finally formulates the dataset in source domain with 2148 denoising (1812 for training and 336 for testing), 2061 dehazing (1500 for training and 561 for testing), 2016 deraining (1568 for training and 448 for testing), 1500 LLIE (1300 for training and 200 for testing) and 890 UIE (700 for training and 190 for testing) images, respectively. All the source domain datasets are applied to training with data augmentation (random flipping, random rotation, and random cropping) to enhance generalization, and all the target domain datasets are applied to testing without any data augmentation. During training, images are randomly cropped to a resolution of 128, and during testing, images are resized to 512 (source domain datasets) and 256 (target domain datasets) resolutions to avoid deployment issues caused by hardware limitations.

\subsection{Evaluation Metrics} \label{sec3.3}
For full-reference datasets (i.e., SIDD, OTS, RealRain-1k, LOL, UIEB, Polyu, and LHP-Rain), the common full-reference metrics PSNR and SSIM \cite{sdir3} are selected to evaluate the IR performance by measuring the discrepancy and similarity between the restored images and their clean reference. For non-reference datasets (including URHI, LIME, and UFO-120), the task-related evaluation metrics are selected to evaluate the IR performance, as summarized as:
\begin{itemize}
	\item For the dehazing task, the FADE \cite{fade} is widely used to estimate haze density in an image by leveraging statistical features related to fog. The FADE has been frequently adopted in dehazing studies, such as in \cite{sdir8}, to provide a non-reference assessment of haze severity.
	\item For the LLIE task, the NIQE \cite{niqe} is a popular non-reference metric that quantifies image quality based on deviations from statistical regularities observed in natural scenes. The NIQE has been extensively employed in LLIE works, such as \cite{sdir4}. In addition, the BRISQUE \cite{brisque} is also frequently used to evaluate perceptual distortions perceptible to the human visual system.
	\item For the UIE task, UIQM \cite{uiqm} and UCIQE \cite{uciqe} are standard evaluation metrics commonly adopted to assess the overall visual quality of enhanced underwater images \cite{sdir7}.
\end{itemize}

\begin{table*}[t!]
	\renewcommand
	\arraystretch{1.5}
	\tabcolsep=0.35cm
	\centering
	\caption{Objective performance comparison of different methods on the source domain datasets, where the \textcolor{red}{best} (\textcolor{blue}{second-best}) performance is indicated in \textcolor{red}{red} (\textcolor{blue}{blue}) font, $\uparrow$ indicates that higher values are better, and $\downarrow$ indicates that lower values are better.}
	\label{tab_sd}
	\resizebox{\linewidth}{!}{
		\begin{tabular}{l|c|cc|cc|cc|cc|cc}
			\hline
			\multirow{2}{*}{\textbf{Method}} & \multirow{2}{*}{\textbf{Venue}} & \multicolumn{2}{c|}{\textbf{Denoising}} & \multicolumn{2}{c|}{\textbf{Dehazing}} & \multicolumn{2}{c|}{\textbf{Deraining}} & \multicolumn{2}{c|}{\textbf{LLIE}} & \multicolumn{2}{c}{\textbf{UIE}} \\
			\cline{3-12}
			 & & SSIM$\uparrow$ & PSNR$\uparrow$ & SSIM$\uparrow$ & PSNR$\uparrow$ & SSIM$\uparrow$ & PSNR$\uparrow$ & SSIM$\uparrow$ & PSNR$\uparrow$ & SSIM$\uparrow$ & PSNR$\uparrow$ \\
			\hline
			AirNet & CVPR'22 & 0.938 & 38.505 & 0.911 & 25.716 & 0.843 & 26.691 & 0.748 & 15.443 & 0.804 & 18.938 \\
            ROP$^+$ & TPAMI'23 & 0.116 & 7.226 & 0.634 & 13.761 & 0.376 & 13.914 & 0.582 & 14.855 & 0.716 & 16.012 \\
			PromptIR & NeurIPS'23 & 0.942 & 38.871 & 0.926 & 27.398 & 0.821 & 27.502 & 0.834 & 19.922 & 0.834 & 21.623 \\
			CAPTNet & TCSVT'24 & 0.913 & 37.27 & 0.932 & 26.301 & 0.686 & 23.327 & 0.709 & 19.226 & 0.821 & 20.423 \\
            DiffUIR & CVPR'24 & \textcolor{blue}{0.950} & \textcolor{blue}{39.777} & 0.930 & 27.703 & 0.872 & 30.053 & \textcolor{blue}{0.867} & 21.279 & \textcolor{red}{0.840} & \textcolor{red}{21.962} \\
            AdaIR & ICLR'25& 0.948 & 39.415 & \textcolor{blue}{0.934} & \textcolor{blue}{28.737} & \textcolor{blue}{0.875} & \textcolor{blue}{30.251} & 0.832 & \textcolor{blue}{21.509} & 0.836 & 21.681 \\
			UDAIR & Ours & \textcolor{red}{0.952} & \textcolor{red}{40.101} & \textcolor{red}{0.938} & \textcolor{red}{29.715} & \textcolor{red}{0.945} & \textcolor{red}{35.193} & \textcolor{red}{0.872} & \textcolor{red}{22.671} & 0.833 & 21.208 \\
			 \hline
		\end{tabular}
	}
\end{table*}

\begin{figure*}[t!]
	\centering
	\includegraphics[width=0.95\linewidth]{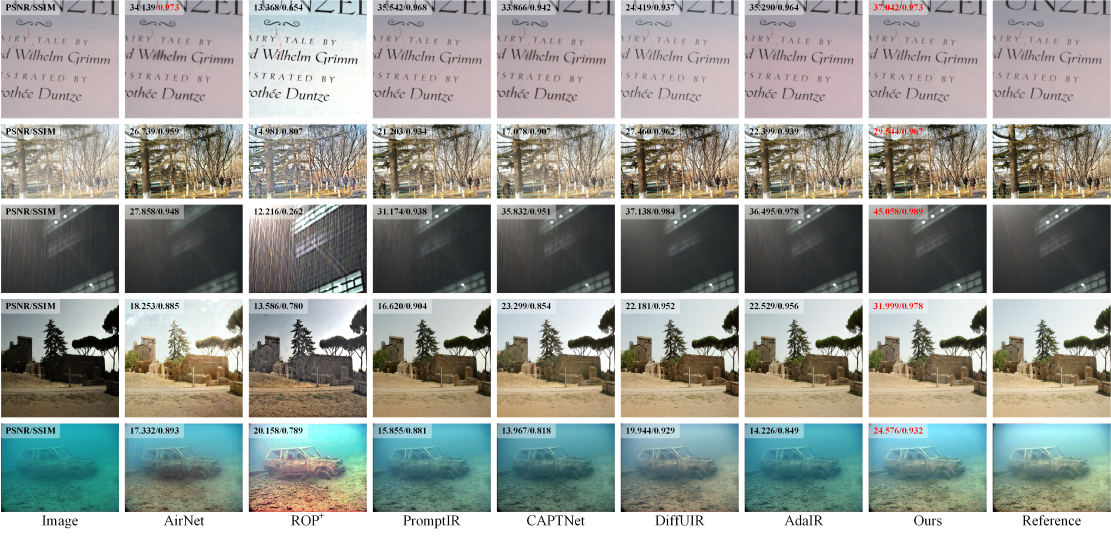}
	\caption{Overall comparisons on the source domain datasets.}
	\label{sd}
\end{figure*}

\begin{figure*}[h]
	\centering
	\includegraphics[width=0.95\linewidth]{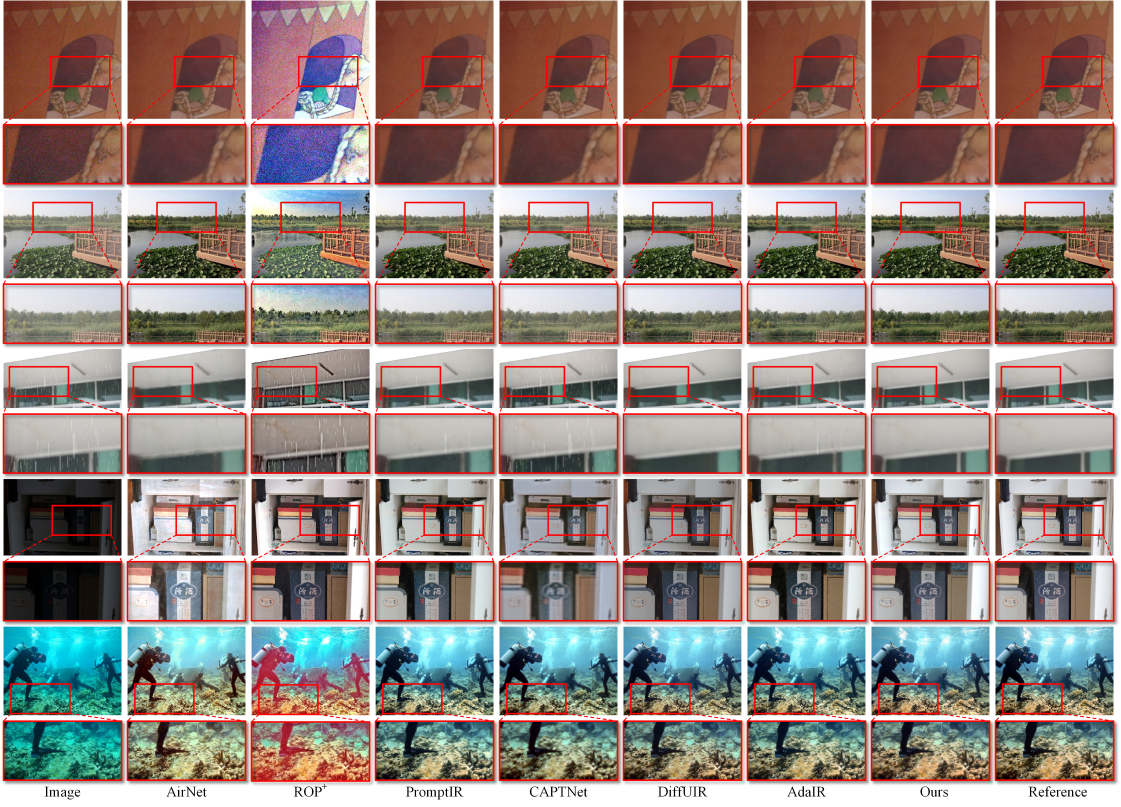}
	\caption{Detail comparisons on the source domain datasets.}
	\label{sd_detail}
\end{figure*}

\subsection{Baselines}
In this work, the following recent AiOIR models are selected as the comparative baselines to comprehensively validate the proposed model, as in:
\begin{itemize}
    \item AirNet \cite{airnet} employs contrastive learning to extract degradation representations from degraded images to guide the network in restoring of clean images.
    \item ROP$^+$ \cite{rop} assumes that degradation can be modeled as a rank-one matrix superimposed on a clean background. ROP$^+$ decomposes the degraded image into a low-rank background (clean image) and a sparse degradation layer (e.g., haze), enabling efficient restoration.
    \item PromptIR \cite{promptir} generates task-specific prompt vectors through a lightweight encoder applied to the input image, dynamically guiding model to adapt to various restoration tasks.
    \item CAPTNet \cite{captnet} decomposes image degradation into independent components (e.g., noise, occlusion) and uses component-oriented prompt vectors to perform targeted restoration.
    \item DiffUIR \cite{diffuir} adopts a diffusion model framework, integrating the progressive denoising process of the diffusion model with the feature selection capabilities of a U-Net-like architecture to iteratively generate clear images.
    \item AdaIR \cite{adair} analyzes the frequency features to separate the frequency sub-bands affected by different degradations (e.g., high-frequency noise, low-frequency haze), enabling targeted modulation of features for restoration.
\end{itemize}  

For all the baselines, default hyperparameter settings are same with the corresponding references to implement the IR tasks.

\subsection{Experimental Configuration} \label{sec3.4}
In this work, to address the issue of imbalanced training samples across tasks,  the samples of each task are randomly resampled with replacement to match the size of the largest dataset. In each iteration, an equal number of samples from all five tasks are randomly selected to form a mini-batch, which is then used for joint training.
The DAAM projects each sample into a 96-dimensional space to better perceive and learn degradation patterns, which subsequently guide the backbone for image restoration. The backbone is configured with a base dimensionality of 24 and progressively processes features by coupling spatial downsampling and upsampling with corresponding changes in feature dimensionality. At the bottleneck layer, the backbone reaches 8 times the base dimensionality with an 8× downsampling rate. Each sample produces 24,576 features under the CSCL constraint. The TTA and DAM are employed during testing time on target domain samples, the step of TTA is set to 5, and DAM adopts the same dimensionality configuration as the decoder to ensure feature compatibility.

All experiments are conducted in the same experimental environment to ensure fairness. The hardware includes an Intel Xeon Gold 5318Y CPU @ 2.10 GHz, 4*NVIDIA GeForce RTX 4090 GPUs and 512 GB of memory. Experiments are performed on Ubuntu 20.04 using the open-source PyTorch 2.4.0 framework. For each baseline, the default hyperparameter are preserved. The proposed method is trained with an AdamW optimizer at an initial learning rate of 1e-4, and cosine annealing schedule is employed to gradually reduce the learning rate throughout the training process.

\subsection{Analysis of Source Domain Results} \label{sec3.5}
The Table \ref{tab_sd} presents a comparison of the objective performance of different methods on the source domain datasets. These results provide insights into how effectively each model learns during training. As shown in Table \ref{tab_sd}, the proposed method achieves the best SSIM and PSNR on denoising, dehazing, deraining, and LLIE, with a particularly large margin on deraining. On the UIE task, our method is slightly weaker but remains close to the top performer. This also demonstrates the strong learning and inference capabilities of the proposed method on the source domain.

Although objective metrics provide a partial indication of model performance, they fail to fully capture the perceptual differences noticed by human observers and typically focus on only a single dimension of quality. Therefore, for image restoration, incorporating subjective visual assessment provides a more comprehensive evaluation. Fig.\ref{sd} and Fig.\ref{sd_detail} show the comparisons on the source domain datasets. From the first to the fifth row, the images correspond to the noisy, hazy, rainy, low‑light, and underwater cases, respectively.

\begin{table*}[t!]
	\renewcommand
	\arraystretch{1.5}
	\tabcolsep=0.35cm
	\centering
	\caption{Objective performance comparison of different methods on the target domain datasets, where the \textcolor{red}{best} (\textcolor{blue}{second-best}) performance is indicated in \textcolor{red}{red} (\textcolor{blue}{blue}) font, $\uparrow$ indicates that higher values are better, and $\downarrow$ indicates that lower values are better.}
	\label{tab_td}
	\resizebox{\linewidth}{!}{
		\begin{tabular}{l|c|cc|cc|cc|cc|cc}
			\hline
			\multirow{2}{*}{\textbf{Method}} & \multirow{2}{*}{\textbf{Venue}} & \multicolumn{2}{c|}{\textbf{Denoising}} & \multicolumn{2}{c|}{\textbf{Dehazing}} & \multicolumn{2}{c|}{\textbf{Deraining}} & \multicolumn{2}{c|}{\textbf{LLIE}} & \multicolumn{2}{c}{\textbf{UIE}} \\
			\cline{3-12}
			 & & SSIM$\uparrow$ & PSNR$\uparrow$ & \multicolumn{2}{c|}{FADE$\downarrow$} & SSIM$\uparrow$ & PSNR$\uparrow$ & NIQE$\downarrow$ & BRISQUE$\downarrow$ & UIQM$\uparrow$ & UCIQE$\uparrow$ \\
			 \hline
			 AirNet & CVPR'22 & 0.744 & 18.426 & \multicolumn{2}{c|}{1.640} & 0.773 & 19.054 & 5.316 & 16.383 & 2.603 & 0.612 \\
             ROP$^+$ & TPAMI'23 & 0.477 & 12.913 & \multicolumn{2}{c|}{\textcolor{red}{0.593}} & 0.556 & 14.718 & 5.290 & 18.196 & \textcolor{red}{3.051} & \textcolor{red}{0.627} \\
			 PromptIR & NeurIPS'23 &	\textcolor{blue}{0.873} & \textcolor{blue}{26.581} & \multicolumn{2}{c|}{1.658} & 0.825 & \textcolor{blue}{21.736} & \textcolor{red}{4.714} & 15.831 & 2.749 & 0.618 \\
			 CAPTNet & TCSVT'24 & 0.783 & 22.211 & \multicolumn{2}{c|}{1.664} & 0.798 & 20.67 & 5.388 & 21.877 & 2.746 & 0.610 \\
             DiffUIR & CVPR'24 & 0.777 & 21.254 & \multicolumn{2}{c|}{1.722} & 0.765 & 20.490 & 6.127 & 15.054 & 2.871 & 0.618 \\
             AdaIR & ICLR'25 & \textcolor{blue}{0.873} & 26.156 & \multicolumn{2}{c|}{1.629} & \textcolor{blue}{0.829} & 21.183 & 4.850 & \textcolor{blue}{15.751} & 2.717 & 0.617 \\
			 UDAIR & Ours & \textcolor{red}{0.883} & \textcolor{red}{27.204} & \multicolumn{2}{c|}{\textcolor{blue}{1.585}} & \textcolor{red}{0.854} & \textcolor{red}{23.765} & \textcolor{blue}{4.814} & \textcolor{red}{13.45} & \textcolor{blue}{2.811} & \textcolor{blue}{0.626} \\
			 \hline
		\end{tabular}
	}
\end{table*}

\begin{figure*}[t!]
	\centering
	\includegraphics[width=0.95\linewidth]{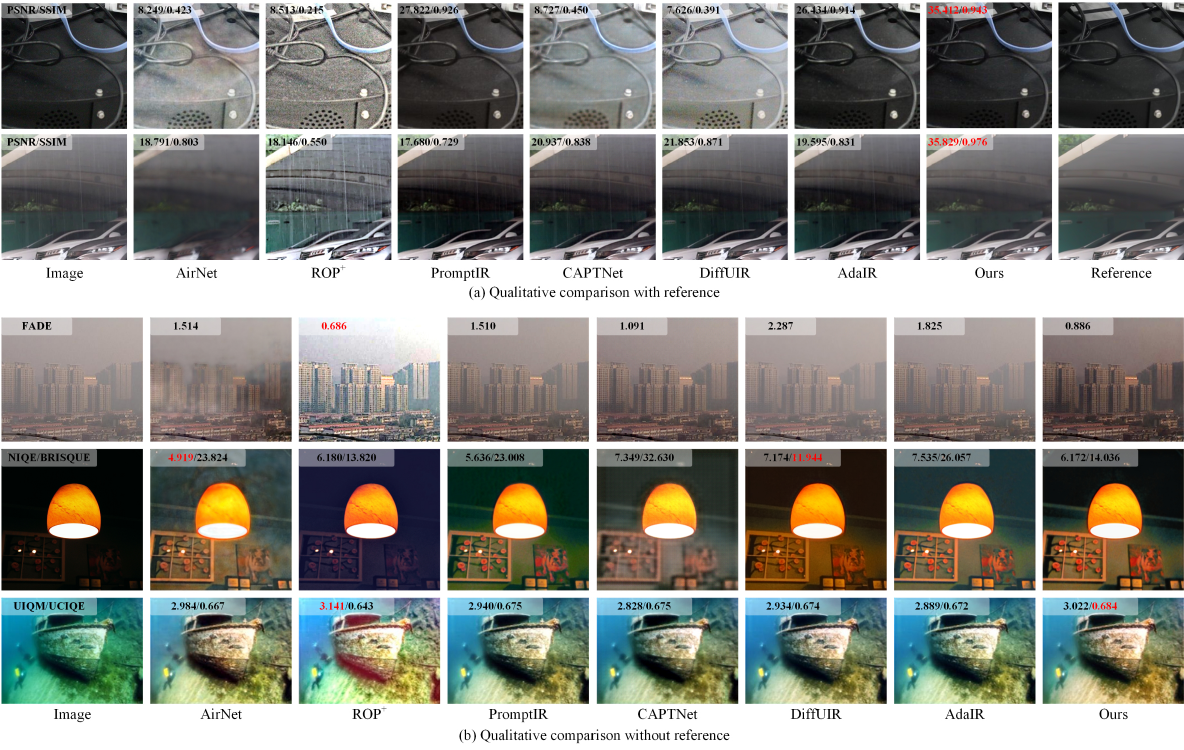}
	\caption{Overall comparisons on the target domain datasets.}
	\label{td}
\end{figure*}

As shown in Fig.\ref{sd}, in the denoising task, AirNet and CAPTNet introduce additional artifacts in text regions, and the text edges appear slightly blurred. ROP$^+$ exhibits noticeable color distortions. In the dehazing task, noticeable haze remains in the results of ROP$^+$, PromptIR, CAPTNet, and AdaIR. In the deraining task, AirNet produces generally unclear results, with residual rain streaks still present. While DiffUIR is capable of removing rain streaks, it suffers from significant loss of texture details. Additionally, ROP$^+$, PromptIR, CAPTNet, and AdaIR also fail to remove the rain effectively. In the LLIE task, AirNet excessively raises the brightness, leading to noticeable overexposure. ROP$^+$, PromptIR and AdaIR still exhibit low brightness in certain regions. CAPTNet achieves a more balanced brightness but produces a blurry image with most texture details missing. In the UIE task, ROP$^+$ and AirNet introduce noticeable artifacts and structural inconsistencies, despite achieving slightly better color fidelity. Meanwhile, PromptIR, CAPTNet, and AdaIR suffer from severe color distortion and significant texture blurring, leading to overall subpar visual quality. Overall, the proposed method effectively suppresses various forms of degradation while accurately restoring fine details and natural colors. Across all tasks, the proposed method consistently produces results that best resemble the reference images.

\begin{table*}[t!]
	\renewcommand
	\arraystretch{1.5}
	\tabcolsep=0.35cm
	\centering
	\caption{Objective performance comparison of different variant in ablation study on the source domain datasets, where the \textcolor{red}{best} (\textcolor{blue}{second-best}) performance is indicated in \textcolor{red}{red} (\textcolor{blue}{blue}) font, $\uparrow$ indicates that higher values are better, and $\downarrow$ indicates that lower values are better.}
	\label{tab_ab_sd}
	\resizebox{\linewidth}{!}{
		\begin{tabular}{l|cc|cc|cc|cc|cc}
			\hline
			\multirow{2}{*}{\textbf{Variant}} & \multicolumn{2}{c|}{\textbf{Denoising}} & \multicolumn{2}{c|}{\textbf{Dehazing}} & \multicolumn{2}{c|}{\textbf{Deraining}} & \multicolumn{2}{c|}{\textbf{LLIE}} & \multicolumn{2}{c}{\textbf{UIE}} \\
			\cline{2-11}
			& SSIM$\uparrow$ & PSNR$\uparrow$ & SSIM$\uparrow$ & PSNR$\uparrow$ & SSIM$\uparrow$ & PSNR$\uparrow$ & SSIM$\uparrow$ & PSNR$\uparrow$ & SSIM$\uparrow$ & PSNR$\uparrow$ \\
			\hline
			baseline & \textcolor{blue}{0.933} & \textcolor{blue}{38.324} & 0.928 & 27.602 & \textcolor{blue}{0.893} & \textcolor{blue}{30.708} & 0.800 & 19.397 & 0.816 & 20.332 \\
			w/o CSCL & 0.929 & 38.021 & \textcolor{blue}{0.929} & \textcolor{blue}{27.669} & 0.870 & 30.161 & \textcolor{blue}{0.824} & \textcolor{blue}{20.045} & \textcolor{red}{0.834} & \textcolor{red}{21.266} \\
			w/o codebook & 0.929 & 37.782 & 0.928 & 27.640 & 0.866 & 30.243 & 0.771 & 18.464 & 0.829 & 20.559 \\
			full mode & \textcolor{red}{0.952} & \textcolor{red}{40.101} & \textcolor{red}{0.938} & \textcolor{red}{29.715} & \textcolor{red}{0.945} & \textcolor{red}{35.193} & \textcolor{red}{0.872} & \textcolor{red}{22.671} & 0.833 & 21.208 \\ 
			\hline
		\end{tabular}
	}
\end{table*}

\begin{table*}[t!]
	\renewcommand
	\arraystretch{1.5}
	\tabcolsep=0.35cm
	\centering
	\caption{Objective performance comparison of different variant in ablation study on the target domain datasets, where the \textcolor{red}{best} (\textcolor{blue}{second-best}) performance is indicated in \textcolor{red}{red} (\textcolor{blue}{blue}) font, $\uparrow$ indicates that higher values are better, and $\downarrow$ indicates that lower values are better.}
	\label{tab_ab_td}
	\resizebox{\linewidth}{!}{
		\begin{tabular}{l|cc|cc|cc|cc|cc}
			\hline
			\multirow{2}{*}{\textbf{Variant}} & \multicolumn{2}{c|}{\textbf{Denoising}} & \multicolumn{2}{c|}{\textbf{Dehazing}} & \multicolumn{2}{c|}{\textbf{Deraining}} & \multicolumn{2}{c|}{\textbf{LLIE}} & \multicolumn{2}{c}{\textbf{UIE}} \\
			\cline{2-11}
			& SSIM$\uparrow$ & PSNR$\uparrow$ & \multicolumn{2}{c|}{FADE$\downarrow$} & SSIM$\uparrow$ & PSNR$\uparrow$ & NIQE$\downarrow$ & BRISQUE$\downarrow$ & UIQM$\uparrow$ & UCIQE$\uparrow$ \\
			\hline
			baseline & 0.841 & 24.992 & \multicolumn{2}{c|}{1.871} & 0.809 & 20.919 & 5.278 & 16.811 & 2.496 & 0.611 \\
			w/o CSCL & \textcolor{blue}{0.872} & \textcolor{blue}{26.836} & \multicolumn{2}{c|}{\textcolor{blue}{1.609}} & \textcolor{blue}{0.844} & \textcolor{blue}{22.784} & 5.062 & 14.929 & 2.692 & 0.618 \\
			w/o codebook & 0.866 & 26.549 & \multicolumn{2}{c|}{1.716} & 0.835 & 21.341 & 5.175 & 15.449 & 2.549 & 0.615 \\
			w/o TTA & 0.871 & 26.768 & \multicolumn{2}{c|}{1.638} & 0.843 & 22.674 & \textcolor{blue}{4.899} & \textcolor{blue}{14.750} & \textcolor{blue}{2.694} & \textcolor{blue}{0.621} \\
			w/o codebook \& TTA & 0.844 & 25.718 & \multicolumn{2}{c|}{1.761} & 0.826 & 21.182 & 5.143 & 15.958 & 2.545 & 0.610 \\
			full mode & \textcolor{red}{0.883} & \textcolor{red}{27.204} & \multicolumn{2}{c|}{\textcolor{red}{1.585}} & \textcolor{red}{0.854} & \textcolor{red}{23.765} & \textcolor{red}{4.814} & \textcolor{red}{13.450} & \textcolor{red}{2.811} & \textcolor{red}{0.626} \\
			\hline
		\end{tabular}
	}
\end{table*}

\begin{figure}[t!]
	\centering
	\includegraphics[width=0.95\linewidth]{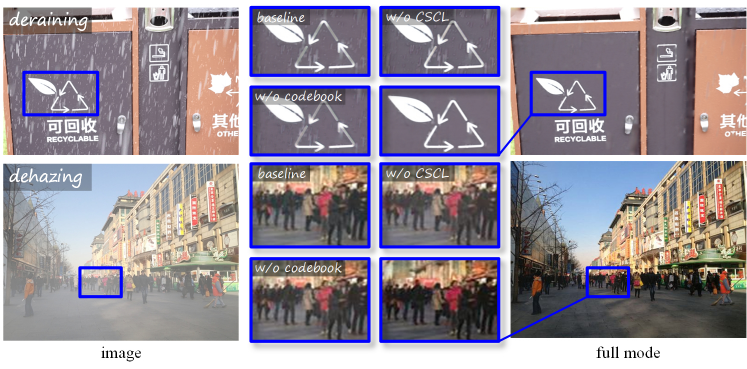}
	\caption{Comparisons of different variants.}
	\label{ab1}
\end{figure}

Fig.\ref{sd_detail} shows the detailed comparisons of different methods. In the denoising task, only AdaIR, DiffUIR, and the proposed method effectively suppress noise in heavily degraded dark regions. For dehazing, AirNet, DiffUIR, and the proposed method are the only models that remove the haze in distant scenes. Although ROP$^+$ also removes haze effectively, it introduces additional artifacts. In the deraining task, ROP$^+$ and CAPTNet leaves rain streaks intact, and PromptIR and AdaIR still retain light streaks. Although AirNet and DiffUIR eliminate the rain, they introduce pronounced blurring artifacts and substantial information loss in results. In the LLIE task, AirNet introduces numerous artifacts, while CAPTNet suffers from significant loss of texture details. In the UIE task, only the proposed method achieves color fidelity close to the reference image. Taken together, the DAAM module and CSCL strategy enable the model to learn and distinguish diverse degradation patterns, allowing the proposed method to excel across all tasks.

\begin{figure*}[t!]
	\centering
	\includegraphics[width=0.95\linewidth]{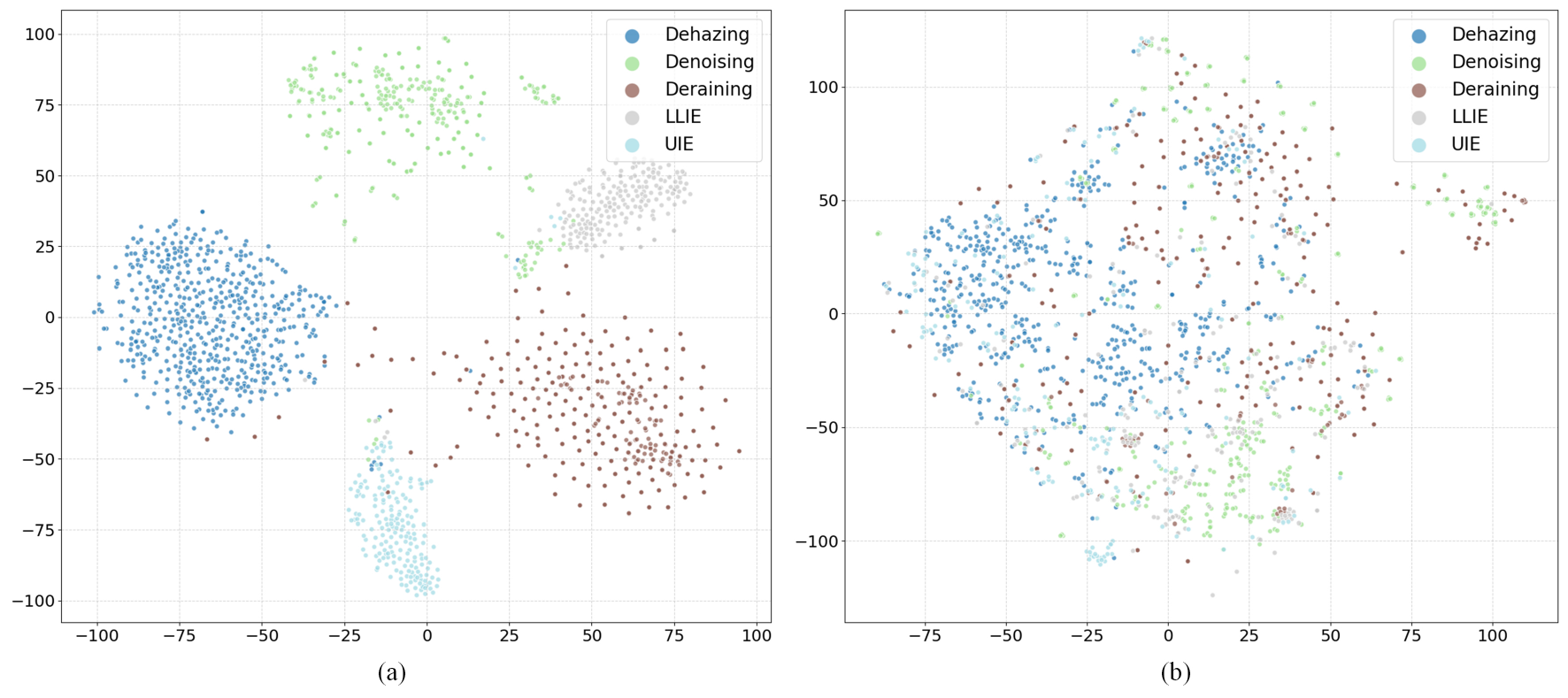}
	\caption{(a) and (b) are t‑SNE visualization of degradation features of w/ CSCL and w/o CSCL.}
	\label{cluster}
\end{figure*}

\begin{figure}[t!]
	\centering
	\includegraphics[width=0.95\linewidth]{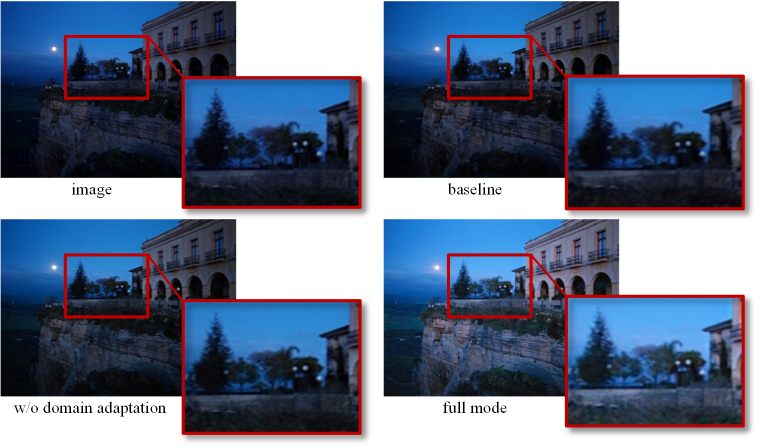}
	\caption{Comparisons of different variants.}
	\label{ab2}
\end{figure}

\subsection{Analysis of Target Domain Results} \label{sec3.6}
Table \ref{tab_td} reports a quantitative comparison of competing methods on the target domain datasets, serving to evaluate the generalization capability of each model. Across the five tasks of denoising, dehazing, deraining, LLIE, and UIE, the proposed method outperforms the baselines in most cases, underscoring the effectiveness of its domain adaptation strategy in cross-domain scenarios. Compared to methods without explicit adaptation mechanisms, our model consistently achieves substantial performance gains on the target domain, indicating that the adaptation scheme significantly improves cross-domain transferability.

More notably, as shown in the subjective visual comparisons in Fig. \ref{td}, the advantages of the proposed method over baselines are even more pronounced. This is primarily because most competing methods struggle to accurately distinguish various degradations in the target domain, resulting in ineffective removal of these degradations. In contrast, our method effectively identifies and suppresses diverse degradations, yielding clearer and more visually pleasing restoration results. Real-world domain shifts are often complex and unpredictable. To this end, the combined use of the codebook and TTA enables dual alignment and dynamic adjustment. Specifically, the codebook achieves inter-domain alignment by establishing a cross-domain latent space that matches degradation pattern prototypes. Meanwhile, TTA ensures intra-domain consistency through embedding refinement, allowing continuous adaptation to target domain variations without extensive annotation or retraining. This synergy allows continuous adaptation to unseen degradations without retraining or annotations on the target domain. It mitigates performance drops caused by data distribution discrepancies and improves both degradation pattern recognition and restoration quality in real-world scenarios.

Fig.\ref{td} presents qualitative results on the target domain datasets: (a) shows the results of denoising and deraining tasks with reference images; (b) shows the results of dehazing, LLIE, and UIE tasks without reference images. In (a), the competing methods struggle with cross‑domain shifts and frequently misidentify degradation patterns. For instance, in the denoising task, ROP$^+$, AirNet, CAPTNet, and DiffUIR mistakenly identify black objects as low-light regions, apply excessive brightness correction, and consequently introduce additional noise and artifacts. In the deraining task, AirNet produces generally blurred results, while ROP$^+$, PromptIR, CAPTNet, DiffUIR, and AdaIR fail to effectively identify and remove rain streaks. By contrast, the proposed method accurately identifies and suppresses both noise and rain while preserving fine textures. (b) shows that only the proposed method reconstructs visually clear images across dehazing, low‑light, and underwater scenarios. AirNet introduces blocky artifacts in all three tasks. Although ROP$^+$ effectively removes haze, it introduces additional noise and exhibits noticeable color casts in both low-light and underwater scenes. PromptIR fails to identify haze-affected scenes and exhibits severe color distortions in both low-light and underwater images. CAPTNet delivers relatively natural colors in the LLIE task but introduces pronounced checkerboard artifacts. DiffUIR fails to effectively identify and remove haze in the dehazing task, while exhibiting color shifts in the LLIE task. AdaIR similarly suffers from color shifts and fails to effectively remove haze. In summary, the proposed method, supported by its domain adaptation strategy, achieves clear image restoration in all five scenarios.

\begin{figure*}[t!]
	\centering
	\includegraphics[width=0.95\linewidth]{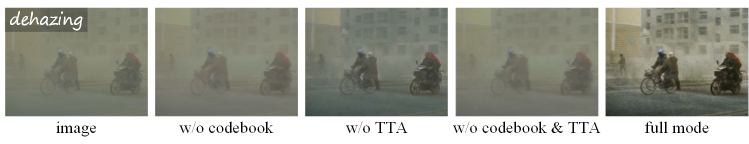}
	\caption{Comparisons of different variants.}
	\label{ab3}
\end{figure*}

\subsection{Ablation Study} \label{sec3.7}
\subsubsection{Degradation Aware and Analysis}
An ablation study is conducted to analyze the role of degradation awareness and analysis within the source domain. Table \ref{tab_ab_sd} reports the quantitative performance of several model variants on the source domain datasets. Fig.\ref{ab1}, Fig.\ref{ab2}, and Fig.\ref{ab3} present qualitative results for the dehazing and deraining tasks. The results show that the two core components of DAAM provide complementary and essential contributions to overall performance improvements. Specifically, the baseline includes only an encoder and decoder, without the proposed DAAM module or any additional strategies. Notably, both CSCL and the domain adaptation strategy depend on the presence of DAAM.

CSCL: Introducing CSCL achieves a great performance improvement across most tasks. CSCL captures both shared and distinctive features across samples with different degradation, enabling the model to learn more separable representations. The features constrained by CSCL guide the backbone to more effectively and precisely remove noise, haze, rain streaks, and other degradations. In addition, the t‑SNE is applied to project the degradation features from DAAM into a two-dimensions space. The resulting visualization, presented in Fig.\ref{cluster} (a) and (b), further demonstrates the pivotal role of CSCL in separating the different degradation categories. Without CSCL, features corresponding to different degradation patterns become heavily mixed. Samples within the same category fail to form clear clusters, indicating the model struggles to distinguish distinct degradation patterns. With CSCL, the five degradation patterns are clustered more compactly and distinctly, with reduced intra-class distances and increased inter-class separations. These results demonstrate that cross‑sample contrastive learning effectively promotes the aggregation of common features within each degradation pattern and the separation of distinct features across patterns. These clearer and more separable representations enable DAAM, together with CSCL, to perceive and distinguish multiple degradations without reference images, thereby providing an effective supervisory signal for restoration.

Codebook: Removing the codebook causes performance decline in every task. This indicates that the discretised codebook not only establishes a cross-domain latent space through effective modality alignment and knowledge distillation but also supplies a stable prior across tasks, yielding robust and generalizable representations for diverse degradations. The full mode achieves the highest evaluation scores across most tasks, confirming that the two designs enable clear restoration of multiple degradations in the source domain.

\subsubsection{Domain Adaptation}
An ablation is performed to study the domain adaptation strategy on the target domain. Table \ref{tab_ab_td} summarizes the objective evaluation results for the different model variants on the target domain datasets. The table shows that the w/o CSCL variant outperforms the baseline, indicating that the domain adaptation mechanism alone provides some degree of alignment and fine‑tuning. However, its performance still falls short of the full model. This suggests that without CSCL, the degradation features are not sufficiently separable, which limits mapping accuracy and subsequent adjustment. The w/o codebook variant shows an even greater performance drop compared to the full mode, suggesting that without the cross-domain latent space, TTA is unable to refine the model based on accurate feature prototypes. While the w/o TTA variant achieves better performance than using CSCL or the codebook alone, it remains inferior to the full mode. This demonstrates that online fine-tuning plays a crucial role in bridging the statistical gap between source and target domains. When both the codebook and TTA are disabled (w/o codebook \& TTA), the metrics nearly revert to baseline levels, confirming that the model struggles with the target domain shift in the absence of explicit alignment and adaptation. In summary, the codebook maps target features into the latent feature space learned from the source domain, TTA refines the model online with respect to target statistics, and CSCL ensures that degradation types are well separated while samples of the same type cluster tightly. Acting together, these three components enable the model to achieve high‑quality image restoration on previously unseen target data.

Fig.\ref{ab2} presents the effect of the domain adaptation strategy in LLIE task. Introducing the domain adaptation strategy (full mode) results in notable visual improvements. 

\begin{figure*}[t!]
	\centering
	\includegraphics[width=0.95\linewidth]{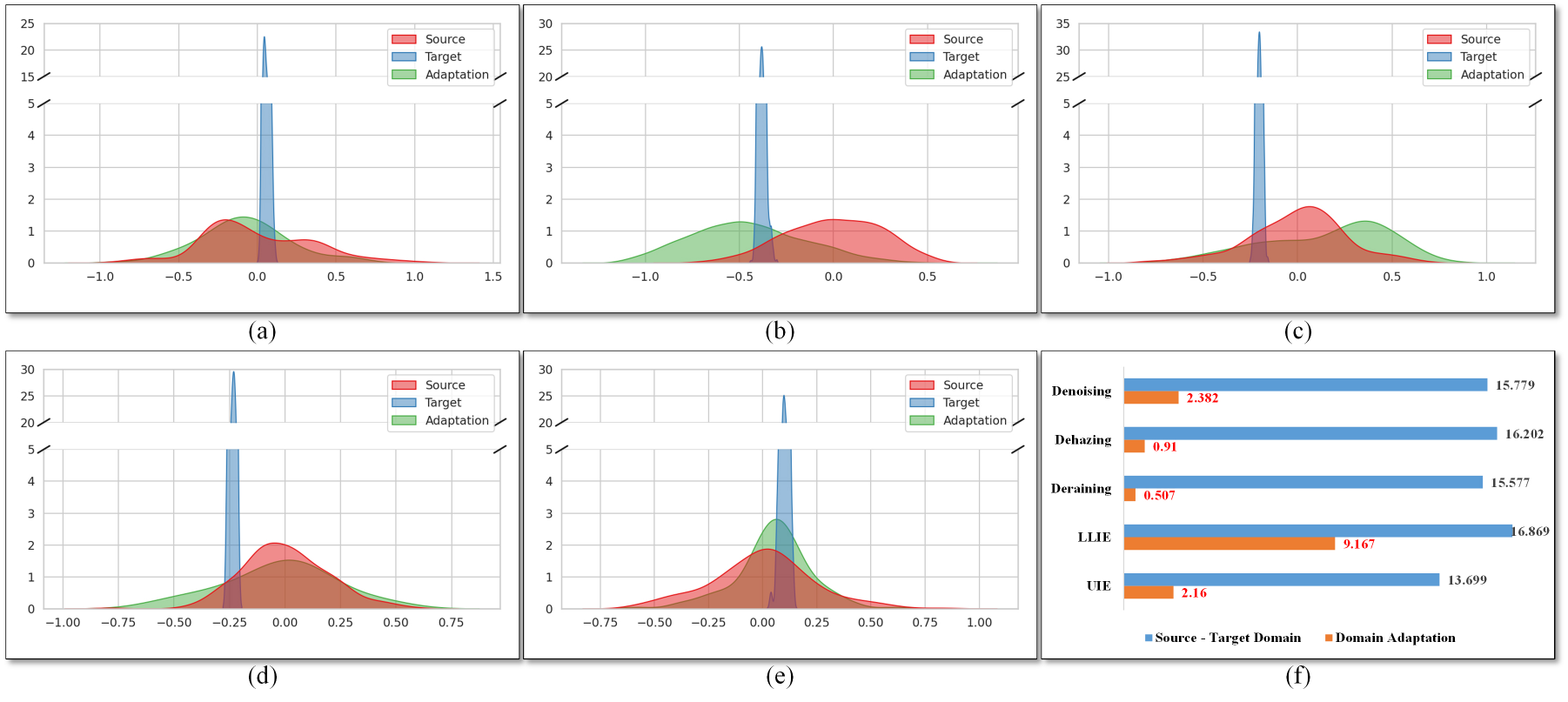}
	\caption{(a) to (e) are density plots analysis in denoising, dehazing, deraining, LLIE, and UIE tasks, respectively. (f) is KL divergence in all five tasks.}
	\label{distribution}
\end{figure*}

In the LLIE task, the baseline and the w/o domain adaptation variant provide only limited brightness increase and lose significant detail in the distant scene. By fusing source domain knowledge and prior with domain adaptation, the full mode variant achieves balanced global brightness while retaining fine textures and natural color contrast. Overall, the domain adaptation strategy maps target domain features into the latent feature space learned from the source domain and refines them during test time. This combination enables more balanced and natural restoration performance across various degradation patterns in the target domain.

Fig.\ref{ab3} illustrates the respective contributions of the codebook and TTA within the domain adaptation strategy. The w/o codebook \& TTA variant (i.e., w/o domain adaptation) appears almost identical to the original hazy image: contrast remains extremely low, and the outlines of distant buildings are obscured. This suggests that the basic restoration architecture alone cannot bridge the distribution gap between the source and target domains. While the codebook offers a cross-domain discrete projection to match degradation pattern prototypes, the absence of sample-wise adaptive correction leads to persistent haze over background structures. On the other hand, using TTA alone proves virtually ineffective, highlighting the necessity of their joint contribution. By contrast, the full mode configuration, which integrates the cross‑domain feature projection via codebook with online fine‑tuning via TTA, eliminates the haze more effectively, demonstrating the clear benefit of combining these two mechanisms.

\subsection{Feature Distribution Analysis} \label{sec3.8}
A experiment is employed to analyze the feature distributions of the source domain, the raw target domain, and the target domain after adaptation. Fig.\ref{distribution} from (a) to (e) show the results of density analysis and KL divergence comparison for data from different domain. The density curves reveal clear differences among the three domains. The source domain (red) exhibits a relatively flat curve with large variance, indicating the features learned in the source domain span a wide range of variations. By contrast, the unadapted target domain (blue) forms a sharp peak with very small variance, showing the extracted features collapse into a narrow interval and cannot represent the diversity of real samples. After domain adaptation (green), the distribution remains relatively concentrated, but its width and position shift closer to those of the source domain. These changes indicate that the domain adaptation strategy successfully mitigates feature distribution discrepancies between domains and enables effective use of the knowledge and priors learned in the source domain to restore target domain images.

\begin{table}[t!]
	\renewcommand
	\arraystretch{1.3}
	\tabcolsep=1cm
	\centering
	\caption{Complexities of different methods.}
	\label{tab3}
	\resizebox{0.95\linewidth}{!}{
		\begin{tabular}{l|cc}
			\hline
			Method & Params. (M) & FLOPs. (G)\\
			\hline
			AirNet & 7.61 & 1209.20 \\
            ROP$^+$ & - & - \\
			PromptIR & 35.59 & 633.62 \\
			CAPTNet & 24.37 & 102.79 \\
            DiffUIR & 36.26 & 398.33 \\
            AdaIR & 28.78 & 589.78 \\
			Ours & 11.24 & 205.64 \\
			\hline
		\end{tabular}
	}
\end{table}

Fig.\ref{distribution} (f) shows the KL divergence of samples from different tasks. The divergence between the source and raw target domains is large, indicating a pronounced distribution shift. After domain adaptation, the divergence between the source and target domains decreases sharply. The substantial decrease in divergence confirms the effectiveness of the adaptation module in narrowing the source–target gap. The method aligns target domain features with representations learned in the source domain, thereby providing more reliable and stable inputs for the subsequent restoration branch.

\subsection{Complexity Analysis} \label{sec3.9}
Table \ref{tab3} reports the parameter counts and computational complexities of the competing methods. Among them, AirNet has the smallest model size with only 7.61M parameters, but suffers from extremely high computational cost, reaching 1209.20G FLOPs, which greatly hinders inference efficiency. PromptIR and AdaIR use 35.59M and 28.78M parameters, respectively, requiring 633.62G and 589.88G FLOPs; DiffUIR also adopts a relatively large architecture with 36.26M parameters and 398.33G FLOPs. These models rely on heavy computation to preserve image quality, but such complexity compromises practical deployment. CAPTNet reduces the computational cost to 102.79G FLOPs with 24.37M parameters, showing high computational efficiency. However, it struggles to restore fine details and suffers from noticeable quality degradation, especially in the presence of complex degradations.

In contrast, the proposed method achieves an excellent balance between performance and efficiency. With only 11.24M parameters and 205.64G FLOPs, it maintains low computational complexity and model size, while nevertheless consistently delivering superior multi-task restoration performance across both source and target domains.

\section{Discussion} \label{sec4}
This work investigates the discrepancy in feature distributions across domains, which remain the primary factor limiting the performance of All-in-One Image Restoration models in diverse real-world scenarios. Such discrepancies also constrain the practical deployment of these models. This issue is expected to be mitigated by introducing a cross-domain latent space mapping and test-time adaptation. A Unified Domain-Adaptive Image Restoration framework is proposed to address two core challenges: handling multiple complex unknown degradations via a unified model architecture, and aligning source-target domain feature distributions through a domain adaptation strategy. Accurate perception and identification of unknown degradations are fundamental to AiOIR. Therefore, the Degradation Aware and Analysis Module is proposed, which employs cross-sample contrastive learning to learn the intrinsic features of degradation pattern without relying on reference images. Within DAAM, the codebook serves dual roles: (1) as a discretizer that quantizes degradation features into latent codes; and (2) as the central adaptation engine. It constructs a cross-domain latent space that store degradation pattern prototypes by learning shared intrinsic priors of degradation pattern, effectively mitigating the impact of performance decline in degradation pattern identification caused by cross-domain distribution shifts. A cross‑sample contrastive learning strategy to prevent the model from focusing on sample‑wise local details that are irrelevant to degradation. By randomly permuting different samples with the same degradation type to form positive and negative pairs, the model is encouraged to capture the intrinsic, common attributes that characterize each degradation. A cross-domain latent space is constructed based on a codebook to store degradation pattern prototypes, aiming to bridge the domain gap. In this way, the model can effectively leverage the knowledge and priors learned from the source domain, alleviating the performance decline in degradation pattern identification caused by distribution shifts. During TTA, the Domain Adaptation Module is dynamically activated to fine-tune the target features through online optimization, thereby minimizing distributional inconsistency. Notably, TTA and DAM are activated exclusively during target domain inference and remains inactive during other cases. This dual mechanism enables the model to effectively leverage source domain degradation priors while adaptively accommodating emerging target domain features, thereby establishing a cross-domain image restoration framework with enhanced robustness and generalization capability. Consequently, the proposed method overcomes the limitations of approaches that rely solely on synthetic or low‑diversity datasets and demonstrates strong generalization in real‑world scenarios. More importantly, the experimental results highlight the essential roles of the domain adaptation strategy, the codebook and TTA mechanism, in generalizing models trained in closed and controlled scenarios to real-world scenarios. This work provides a solid foundation for exploring more effective techniques for real-world applications in the future.

\textbf{Future Directions.} Current AiOIR implementations focus on automatically identifying and restoring single degradation patterns. However, this approach differs significantly from practical scenarios where composite degradations, such as simultaneous rain, haze, and low-light conditions, frequently coexist. This capability gap stems from the prevailing practice of training on single degradation datasets, which restricts models to sequential restoration rather than holistic restoration. The imperative need for multi-degradation benchmark datasets is emphasized to address this critical challenge. Such datasets systematically simulate real-world degradation combinations, enabling AiOIR methods to simultaneously recognize heterogeneous degradations and holistically restore clear images through unified frameworks. This direction holds paramount significance for advancing AiOIR research while unlocking substantive practical value for real-world vision systems operating under complex environmental conditions.

\section{Acknowledgements}
This work was supported by National Natural Science Foundation of China (NSFC) 62371323, U2433217, U2333209, Sichuan Science and Technology Program, China 2024YFG0010, 2024ZDZX0046, and Institutional Research Fund from Sichuan University 2024SCUQJTX030.

{\small
\bibliographystyle{IEEEtran}
% \bibliography{refs}

}

\end{document}